\documentclass[sigconf]{acmart}
\usepackage{graphicx}
\usepackage{multirow}
 \usepackage{enumitem}

\AtBeginDocument{%
  \providecommand\BibTeX{{%
    \normalfont B\kern-0.5em{\scshape i\kern-0.25em b}\kern-0.8em\TeX}}}

\setcopyright{acmcopyright}
\copyrightyear{2024}
\acmYear{2024}
\acmDOI{XXXXXXX.XXXXXXX}

\acmPrice{15.00}
\acmISBN{978-1-4503-XXXX-X/18/06}

\begin{document}

\title{Learning Multi-graph Structure for Temporal Knowledge Graph Reasoning}

\author{Jinchuan Zhang}

\affiliation{%
  \institution{School of Computer Science and Engineering, University of Electronic Science and Technology of China}
  \city{Chengdu}
  \state{Sichuan}
  \country{China}
}
\email{jinchuanz@std.uestc.edu.cn}

\author{Bei Hui}
\affiliation{%
  \institution{School of Information and Software Engineering, University of Electronic Science and Technology of China}
  \city{Chengdu}
  \state{Sichuan}
  \country{China}
}
\email{bhui@uestc.edu.cn}

\author{Chong Mu}
\affiliation{%
  \institution{School of Information and Software Engineering, University of Electronic Science and Technology of China}
  \city{Chengdu}
  \state{Sichuan}
  \country{China}
}
\email{muchong@std.uestc.edu.cn}

\author{Ling Tian}
\affiliation{%
  \institution{School of Computer Science and Engineering, University of Electronic Science and Technology of China}
  \city{Chengdu}
  \state{Sichuan}
  \country{China}
}
\email{lingtian@uestc.edu.cn}





\renewcommand{\shortauthors}{Jinchuan, et al.}

\begin{abstract}
Temporal Knowledge Graph (TKG) reasoning that forecasts future events based on historical snapshots distributed over timestamps is denoted as extrapolation and has gained significant attention.
Owing to its extreme versatility and variation in spatial and temporal correlations,
TKG reasoning presents a challenging task,
demanding efficient capture of concurrent structures and evolutional interactions among facts.
While existing methods have made strides in this direction, they still fall short of harnessing the diverse forms of intrinsic expressive semantics of TKGs, which encompass entity correlations across multiple timestamps and periodicity of temporal information.
This limitation constrains their ability to thoroughly reflect historical dependencies and future trends. 
In response to these drawbacks, this paper proposes an innovative reasoning approach that focuses on \underline{L}earning \underline{M}ulti-graph \underline{S}tructure ($\mathsf{LMS}$).
Concretely, it comprises three distinct modules concentrating on multiple aspects of graph structure knowledge within TKGs, including concurrent and evolutional patterns along timestamps, query-specific correlations across timestamps, and semantic dependencies of timestamps, which capture TKG features from various perspectives. 
Besides, $\mathsf{LMS}$ incorporates an adaptive gate for merging entity representations both along and across timestamps effectively. 
Moreover, it integrates timestamp semantics into graph attention calculations and time-aware decoders, in order to impose temporal constraints on events and narrow down prediction scopes with historical statistics. 
Extensive experimental results on five event-based benchmark datasets demonstrate that $\mathsf{LMS}$ outperforms state-of-the-art extrapolation models, indicating the superiority of modeling a multi-graph perspective for TKG reasoning.
\end{abstract}





\maketitle

\section{Introduction}
Temporal Knowledge Graphs (TKGs) \cite{tkgsurvey} have emerged as a powerful solution to address the limitations of traditional static KGs in capturing the dynamic evolution of real-world events. It incorporates temporal information to enhance dynamic events represented as (\textit{subject}, \textit{relation}, \textit{object}, \textit{timestamp}) quadruples, effectively bridging the gap between the static KGs and the dynamic characteristics. Meanwhile, this innovation of temporal enrichment has significantly expanded the application domains for KGs, encompassing event prediction \cite{zhouwt}, social network analysis \cite{snet}, sensor fault detection \cite{Iot}, \textit{etc}.

In general, TKGs can be conceptualized as comprehensive multi-relational graphs, similar in structure to static KGs but enriched with timestamped relation edges to convey dynamic characteristics. To better represent the evolutional patterns, TKGs also can be formulated as a sequence of snapshots or subgraphs partitioned according to time intervals, where each snapshot encapsulates all events that occurred at a specific timestamp, providing a more intuitive view of temporal evolution.
Nevertheless, owing to its inherent temporal nature, TKGs face two unique incompleteness challenges: interpolation and extrapolation, which pertain to inferring information at past and future timestamps, respectively. It is worth noting that this paper primarily focuses on the task of extrapolation in TKG reasoning, while its principles and methodologies can be adapted to handle interpolation scenarios.
\begin{figure}[h]
\centerline{\includegraphics[width=1\linewidth]{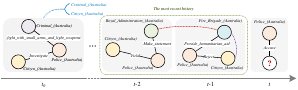}}
\caption{An illustration of extrapolation over TKGs.}
\label{ill-tkg}
\end{figure}

To comprehensively explore the rich semantics within TKGs and accurately predict future events, it is essential to leverage the entire TKG and its diverse interactions effectively. An example of ICEWS \cite{icews} dataset is illustrated in Figure \ref{ill-tkg}, TKGs encompass various types of dependencies that offer valuable and distinct information for extrapolation task, including global insights derived from a distant historical timestamp (\textit{e.g.}, $t_0$ highlighted by the blue dashed edge), and structural and evolutional patterns of the most recent events (depicted within the green dotted box).
Significantly, the global insights provide relevant repetitive entities that ever occurred to queries, while the structural interactions among neighboring entities encompass connections both across timestamps and within concurrent facts (indicated by red and purple dashed edges, respectively), both of which play a vital role in predicting potential outcomes and future existences. 

Several earlier works \cite{cygnet,xerte,titer} introduce novel models that emphasize different aspects of TKG reasoning, including historical repetitive statistics \cite{cygnet}, subgraph search strategies \cite{xerte}, and the incorporation of reinforcement learning \cite{titer}.
Furthermore, a series of preceding extrapolation approaches \cite{renet,regcn,evokg,cen,tirgn,hgls,rpc} have made efforts to leverage Graph Neural Networks (GNNs) or Graph Attention Networks (GATs) to model the structural dependencies among concurrent facts, and utilize recurrent units to capture the sequential evolution of TKGs. Within this series, TiRGN \cite{tirgn} incorporates global historical context to constrain the prediction scope, while HGLS \cite{hgls} establishes connections between common entities across different timestamps and designs a hierarchical GNN to capture multi-hop correlations. RPC \cite{rpc} constructs a relational correlation graph to extract correspondences and introduces a weight assignment mechanism for temporal information from snapshots. However, these methods still exhibit certain limitations in terms of holistically capturing the diverse dependencies present in TKGs, despite their promising performances.

In one aspect, a majority of existing methods lack effective and explicit handling of entity correlations across different snapshots. This deficiency arouses the loss of crucial dependencies between recent and past events that could serve as potential valuable clues for TKG reasoning. Another noteworthy aspect is that prior approaches tend to neglect valuable information inherent in timestamps, such as natural periodic semantics and interactions, giving rise to inadequate representations. Indeed, only a few insights have been proposed to address the former constraint. For instance, HGLS \cite{hgls} introduces a global graph that connects each snapshot through additional edges to capture long-term entity dependencies. Nevertheless, it may incorporate redundant information from distant timestamps because of the excessively long historical length, which also necessitates more hops and layers for precisely modeling neighboring interactions, leading to unnecessary computational overhead. Despite an appropriate historical length, an abundance of facts can impede the capacity to capture entity characteristics across different timestamps. This emphasizes the need to reduce excessive connections in recent snapshots, which is also essential for capturing key correlations.

To tackle these aforementioned challenges, this paper proposes a novel method that specializes in \underline{L}earning \underline{M}ulti-graph \underline{S}tructure ($\mathsf{LMS}$) from various perspectives of TKG semantics (i.e., evolutional graph, union graph, and temporal graph) for effective reasoning. 
Specifically, $\mathsf{LMS}$ begins by focusing on the most recent history, capturing concurrent interactions and temporal patterns through the Evolutional Graph Learning (EGL) (\S \ref{egl}). To explicitly model structural correlations among crucial entities across timestamps, $\mathsf{LMS}$ further introduces Union Graph Learning (UGL) (\S \ref{ugl}), which constructs a union graph that focuses on query subjects, emphasizing events most relevant to the future. Additionally, from a global perspective, we present Temporal Graph Learning (TGL) (\S \ref{tgl}) to capture semantics and periodic connections between timestamps via a graph-based approach.
Furthermore, $\mathsf{LMS}$ incorporates an indicator (\S \ref{ind}) within time-aware decoders (\S \ref{dec}) to make statistics of whole historical repetitive events and thereby narrow down the scope of predictions for one decoder. The above achieves entity prediction by considering both realistic occurrence and historical context in TKGs.
In general, the main contributions of this paper can be summarized as follows:
\begin{itemize}[leftmargin=*]
    \item We propose a superior temporal reasoning model named $\mathsf{LMS}$, which sufficiently learns various structural and temporal dependencies in TKGs from a multi-graph perspective. To the best of our knowledge, $\mathsf{LMS}$ is the first study that constructs a query-specific union graph to learn structural correlations across time and integrates temporal information into a graph view for TKG extrapolation.
    \item Through the evolutional, union, and temporal graph learning, $\mathsf{LMS}$ is able to holistically integrate TKG representations with concurrent and sequential interactions of entities and temporal semantic information.
    \item Extensive experiments on five benchmarks demonstrate the learning ability of $\mathsf{LMS}$ for relevant information contained in TKGs, as well as its effectiveness and significant performance (up to 2.60\% and 6.47\% improvement in time-aware filtered and raw MRR) compared to state-of-the-art methods.
\end{itemize}
\section{Related Works}
\begin{figure*}[]
\centerline{\includegraphics[width=1\linewidth]{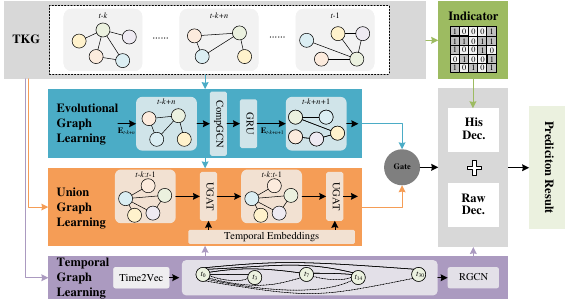}}
\caption{An illustration of $\mathsf{LMS}$ model architecture. Firstly, a series of TKG sequences are learned through \textit{Evolutional Graph Learning} (\S \ref{egl}). Then, \textit{Union Graph Learning} (\S \ref{ugl}) constructs a query-specific graph based on the sequence and queries and captures the structural interactions among those facts, while temporal information (\S \ref{tgl}) is combined in this step. After that, the \textit{Time-aware Decoder} (\S \ref{dec}) is adopted to predict future facts according to \textit{Indicator} (\S \ref{ind}) and the combined features from \textit{Adaptive Gate} (\S \ref{gate}).}
\label{lms_framework}
\end{figure*}
\subsection{Static Knowledge Graph Reasoning}
Early efforts in KG reasoning focused on static KGs \cite{kgsurvey}. DistMult \cite{distmult} learns relational semantics via embeddings from a bilinear objective. ComplEx \cite{complex} employs complex embeddings and Hermitian dot products. RotatE \cite{rotate} represents relations as rotations in complex vector space. With the evolution of KG representation learning, researchers emphasized the importance of deep learning and modeling relations. Methods like ConvE \cite{conve} and ConvTransE \cite{convtranse} utilize convolutional operations for entity-relation embedding fusion. Graph Convolutional Networks (GCN) \cite{gcn} learn KG structural features, with R-GCN\cite{rgcn} extending GCNs with relation-specific block-diagonal matrices. And CompGCN\cite{compgcn} integrates multi-relational GCN methods and various composition operations, including translation assumption $s+r\approx o$ of TransE \cite{transe}. Nevertheless, these approaches often struggle to incorporate temporal information, rendering them less effective in handling dynamic knowledge graphs in practical scenarios.
\subsection{Temporal Knowledge Graph Reasoning}
According to the timestamps of predicted events, such as the occurred moment or future, TKG reasoning can be roughly divided into two types: interpolation and extrapolation.
\subsubsection{Interpolation setting}
Similar to KG completion, the interpolation setting aims to infer missing facts at specific past timestamps. Early approaches tackle this by incorporating temporal information within the realm of static KG reasoning. Notable methods include tTransE \cite{ttranse} and TA-DistMult \cite{tadistmult}, which introduce temporal order constraints to ensure the consistency of embeddings over time. ConT \cite{ConT} extends static models to episodic tensors, enabling the incorporation of semantic and episodic memory \cite{ypmDM} functions. HyTE \cite{hyte} constructs a hyperplane for each timestamp to represent the facts at that particular TKG snapshot. Inspired by diachronic word embeddings, DE-SimplE \cite{desimple} introduces a diachronic embedding function to capture entity features at various timestamps. Additionally, TNTComplEX \cite{tntcomplex} and ChronoR \cite{chronor} extend existing models, ComplEX \cite{complex} and RotatE \cite{rotate}, respectively, to address temporal reasoning within KGs. Since many existing models are based on Euclidean space, DyERNIE \cite{dyernie} learns evolving entity representations of TKGs through a product of Riemannian manifolds, enabling a more comprehensive reflection of diverse geometric structures and evolutionary dynamics. In order to summarize and compare the above interpolation methods, \cite{EvaTKGs} provides a unified open-source composability framework that can arbitrarily combine any part of those models.
\subsubsection{Extrapolation setting}
The extrapolation task aims to predict future facts at timestamps $t>t_T$ based on historical TKG sequences, a variety of approaches have been explored in recent years. Know-Evolve \cite{kevolve}, GHNN \cite{ghnn}, and DyREP \cite{dyrep} utilize temporal point processes like the Rayleigh process and Hawkes process to model TKG evolution. CyGNet \cite{cygnet} employs a copy-generation mechanism and multi-hot vectors to construct a historical vocabulary indicating entities from the past. xERTE \cite{xerte} designs a subgraph sampling module based on temporal information. TANGO \cite{tango} extends neural ordinary equations (ODEs) to learn continuous-time representations and introduces a graph transition layer to enhance predictions by focusing on neighboring observations. CluSTER \cite{cluster} and TITer \cite{titer} adopt reinforcement learning to generate query-specific paths comprised of historical entities. TLogic \cite{TLogic} introduces an explainable forecasting framework based on temporal logical rules within TKGs. CEN \cite{cen} models the complex TKG evolution by considering historical length-diversity and time-variability.

Besides, RE-NET \cite{renet}, RE-GCN \cite{regcn}, TiRGN \cite{tirgn}, HGLS \cite{hgls}, and RPC \cite{rpc} are the most relevant works to $\mathsf{LMS}$, which employ graph neural networks to model concurrent structures and utilize recurrent units to learn entity or relation evolution patterns. Specifically, RE-NET \cite{renet} only focuses on the query level and 1-hop neighbors, while RE-GCN \cite{regcn} treats each snapshot of the TKG sequence as a whole input and introduces static constraints to initialize entity embeddings. TiRGN \cite{tirgn} adopts a global history encoder to identify entities related to historical repetitive facts and a time-guided decoder that incorporates timestamps for predictions. HGLS \cite{hgls} explicitly captures long-term time dependencies and adaptively combines long- and short-term information for entity prediction. RPC \cite{rpc} extracts graph structural information and temporal interactions through relational correlation and periodic pattern correspondence units, separately.

Above all, none of these methods explicitly capture the multi-graph structural dependencies from TKGs, such as the query-specific entities that are directly related to future events, and the realistic semantic interactions between timestamps.

\section{Method}
The overarching framework of $\mathsf{LMS}$ is depicted in Figure \ref{lms_framework}, which consists of five key components: (1) \textit{Evolutional Graph Learning}, which focuses on capturing correlations among concurrent facts and modeling temporal evolution of snapshots; (2) \textit{Union Graph Learning} which constructs a query-specific union graph based on snapshots and learns the structural information within this graph; (3) \textit{Temporal Graph Learning}, which constructs an auxiliary graph that incorporates timestamps and models the periodic interaction between them; (4) \textit{Indicator} is responsible for gathering statistics related to historical facts and using them to refine predictions; (5) \textit{Time-aware Decoder} that integrates temporal information in ConvTransE to predict future facts.

\subsection{Notations}
In this study, Temporal Knowledge Graphs (TKGs) are represented as sequences of subgraphs, referred to as snapshots, formalized as $G = \{G_1, G_2, ..., G_\mathcal{T}\}$. Here $\mathcal{T}$ denotes the total number of timestamps, and each $G_t$ corresponds to the collection of facts or events occurring at timestamp $t$.
Given $G=\{\mathcal{E}, \mathcal{R}, \mathcal{F}, \mathcal{T}\}$, the $\mathcal{E}$, $\mathcal{R}$, and $\mathcal{F}$ denote the sets of entities, relations, and facts, respectively. Each fact $f \in \mathcal{F}$ is defined as a quadruple $(s, r, o, t)$, where $s$ and $o$ belong to $\mathcal{E}$, representing the subject and object entities respectively, $r \in \mathcal{R}$ represents the relation, and $t \in \mathcal{T}$ indicates the timestamp of the quadruple. Throughout the paper, bold letters indicate vectors.

Moreover, the extrapolation task for TKGs falls under the category of link prediction in future timestamps. Given a query $q = (s, r, ?, t)$, with $q \in \mathcal{Q}$, the objective is to compute the conditional probability of an object $o$ given the subject $s$, relation $r$, timestamp $t$, and the past snapshots, denoted as $p(o|s, r, t, G_{t-1})$.
\subsection{Evolution Graph Learning} \label{egl}
The fundamental principle of extrapolation involves forecasting forthcoming events based on historical snapshots from the immediate past. In line with this concept, we regard neighboring $k$ snapshots as histories and aggregates features of those snapshots from both structural and temporal aspects, defined as Evolution Graph Learning (EGL).
Specifically, in the structural aspect, the basic aggregation process bears resemblance to that of CompGCN \cite{compgcn}. To facilitate the merging of entities and relations, the composition operator is substituted with a one-dimensional convolution, similar to TiRGN \cite{tirgn}. The formulation is as follows:
\begin{equation}
    \textbf{o}^{(l+1)}_t=\sigma(\sum_{(s,r,o)\in \mathcal{F}_t}\textbf{W}^{(l)}_1(\psi(\textbf{s}^{(l)}_t ||\textbf{r}_t))+\textbf{W}^{(l)}_2\textbf{o}^{(l)}_t)\textnormal{,}
\end{equation}
where $\sigma$ represents the Randomized leaky Rectified Linear Unit (RReLU) activation function, $\psi$ stands for the one-dimensional convolution operator, $||$ denotes the concatenation, $\mathbf{W}^{(l)}_1$ and $\mathbf{W}^{(l)}_2$ are the trainable weight matrices for $(s, r)$ and the self-loop of objects in this CompGCN layer. The resulting entity representation following aggregation for each snapshot is denoted as $\mathbf{G}_t$.

In the temporal aspect, the Gated Recurrent Unit (GRU) is employed to progressively capture evolving entity and relation representations. The updated embeddings $\textbf{E}_t$ for entities can be expressed as:
\begin{equation}
    \textbf{E}_t = \mathrm{GRU}(\textbf{E}_{t-1},\textbf{G}_{t-1})\textnormal{.}
\end{equation}
Subsequently, all the outputs of GRU can be organized into a list $\{\textbf{E}_{t-k+1},\textbf{E}_{t-k+2},...,\textbf{E}_{t}\}$ for further learning. Similarly, the relation embeddings $\textbf{R}_t$ are updated as follows:
\begin{equation}
    \textbf{R}_t = \mathrm{GRU}(\textbf{R}_{t-1},[pooling(\textbf{E}^{\mathcal{R}}_{t-1})||\textbf{R}])\textnormal{,}
\end{equation}
where \textit{pooling} denotes the mean pooling operation, $\textbf{E}^\mathcal{R}$ denotes the embeddings of $\mathcal{R}$ related entities, $\textbf{R}$ is the initial embeddings of relations.
\subsection{Union Graph Learning} \label{ugl}
While EGL adeptly captures evolutional trends, it falls short in spotlighting pivotal events closely related to queries, which can significantly impact predictions and deserve special consideration. Moreover, it is crucial to amalgamate these events into a union graph enriched with temporal attributes, which can leverage a comprehensive historical foundation for extrapolation and explicit modeling of correlations and temporal effects. In light of these considerations, we present query-specific Union Graph Learning (UGL) that creates a union graph comprising entities specifically relevant to given queries, along with relations annotated with temporal information. The primary aim of UGL is to capture distinctive correlations and temporal interactions among these entities. 
\subsubsection{Construction and Aggregation}
\begin{figure}[]
\centerline{\includegraphics[width=1\linewidth]{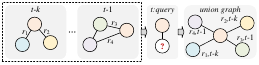}}
\caption{An illustration of the union graph construction.}
\label{ugl-constru}
\end{figure}

Our approach involves the extraction of all related facts, denoted as $f_q \in \{G_{t-k:t-1}\}$, from historical snapshots in accordance with the query set $\mathcal{Q}_t$, where $q=(s,r,?,t)$ and $f_q$ must contain $s$ as the subject or object. Subsequently, we construct a union graph, denoted as $UG$, using these query-specific facts. This process can be viewed as Figure \ref{ugl-constru}. 
Importantly, we retain temporal information by incorporating the original timestamp $t$, as an attribute of the relations within TKGs. This augmentation allows UGL to effectively characterize timestamp-related features, providing a richer representation of facts at different timestamps within the union graph.

First of all, we use the mean of the sequence from EGL as the initial embeddings $\textbf{UE}_{t-k:t-1}$ of UGL, which is represented as:
\begin{equation}
    \textbf{UE}_{t-k:t-1}=\mathrm{MEAN}(\sum_{i = t-k+1}^{t} \textbf{E}_i)\textnormal{.}
\end{equation}
Then the Union GAT (UGAT), which integrates temporal information, is applied to obtain the attention scores $\alpha_{o,s}$ as follows:
\begin{equation}
    \alpha^{(l)}_{o,s} = \frac{exp(\textbf{W}^{(l)}_3\sigma(\textbf{W}^{(l)}_4[\textbf{s}^{(l)}||\textbf{r}||\textbf{o}^{(l)}||\textbf{t}^{\prime \prime}]))}{\sum_{s' \in \mathcal{N}_{(o)}}exp(\textbf{W}^{(l)}_3\sigma(\textbf{W}^{(l)}_4[\textbf{s'}^{(l)}||\textbf{r}||\textbf{o}^{(l)}||\textbf{t}^{\prime \prime}]))}\textnormal{,}
\end{equation}
where $\textbf{s}, \textbf{o} \in \textbf{UE}_{t-k:t-1}$, $\textbf{W}^{(l)}_3\in\mathbb{R}^{4d}$ and $\textbf{W}^{(l)}_4\in\mathbb{R}^{4d \times 4d}$ represent linear transformation layers, $\sigma$ corresponds to the LeakyReLU activation function, $\mathcal{N}_{(o)}$ denotes the set of subjects for entity $o$. Subsequently, the aggregator of UGL is defined as follows:
\begin{equation}
    \textbf{o}^{(l+1)}=\sigma(\sum_{(s,r,o)\in UG} \alpha^{(l)}_{o,s}\textbf{W}^{(l)}_5\psi(\textbf{s}^{(l)}||\textbf{r})+\textbf{W}^{(l)}_6\textbf{o}^{(l)}) \textnormal{,}
\end{equation}
where $\sigma$ is the RReLU activation function, $\textbf{W}^{(l)}_5$ and $\textbf{W}^{(l)}_6$ are weight matrices. The output of UGL can be represented as $\textbf{UE}_t$.
\subsubsection{Adaptive Gate} \label{gate}
Following the aggregation process through UGL, it becomes imperative for $\mathsf{LMS}$ to adaptively combine the features from both the evolution and union of TKGs. To achieve this, we employ a learnable adaptive gate, denoted as $\Theta \in \mathbb{R}^{|\mathcal{E}| \times d}$, which serves to regulate the weighting of the evolutional and union representations. Formally,
\begin{equation}
    \textbf{GE}_t = \sigma(\textbf{W}_7\Theta_e)\textbf{E}_t + (1 - \sigma(\textbf{W}_7\Theta_e))\textbf{UE}_t \textnormal{,}
\end{equation}
where $\sigma(\cdot)$ is the \textit{Sigmoid} activation function to limit the weight of evolution between $[0, 1]$, $\textbf{W}_7\in\mathbb{R}^{1 \times d}$ is the linear transformation.
\subsection{Temporal Graph Learning} \label{tgl}
Most existing approaches lack the capability to capture meaningful correlations between timestamps, limiting their ability to model natural temporal periodicity and variability effectively. 
In practice, numerous events follow a recurring pattern, such as occurring every Monday or on the first day of each month. Addressing this issue, we regard time information as one of the basic semantic attributes of TKGs and propose Temporal Graph Learning (TGL) with empirical periodicity to establish an intuitive connection between those temporal nodes.
A part of the Temporal Graph in $\mathsf{LMS}$ is illustrated in Figure \ref{lms_framework}, where nodes of timestamps are connected through relations and corresponding inverse (dash arrow) based on realistic granularity to capture the periodic characteristics of different granularities and impose constraints on the nodes. 
For instance, the "one week later" relation links temporal nodes that denote the same day of the week. The definition of temporal relations in ICEWS datasets is shown in Table \ref{t_relation}, where the timestamps in GDELT are connected in the "hour" dimension.
\begin{table}[h]
\small
\caption{Definitions of relations in Temporal Graph.}
\begin{tabular}{ccc}
\hline
Types & Relation & inverse\_Relation \\ \hline
0 & three days later & three days ago \\
1 & one week later & one week ago \\
2 & two weeks later & two weeks ago \\
3 & one month later & one month ago \\ \hline
\label{t_relation}
\end{tabular}%
\end{table}

The temporal graph is denoted as $TG=\{\mathcal{E}_{TG}, \mathcal{R}_{TG}\}$, where $\mathcal{E}_{TG}$ determined according to the number of timestamps $\mathcal{T}$, and relations in $\mathcal{R}_{TG}$ representing time periods of various granularities.
To obtain representations for the temporal graph, we initiate embeddings $\textbf{t}$ based on Time2Vec\citep{t2v}:
\begin{equation}
    \textbf{t}^{\prime} = \textbf{W}_8(\textbf{W}_{9}\textbf{t}||\sigma(\textbf{W}_{10}\textbf{t})) \textnormal{,}
\end{equation}
where $\sigma$ denotes Sine activation function, $\textbf{W}_8\in\mathbb{R}^{d \times 2d}$, $\textbf{W}_{9}\in\mathbb{R}^{d \times 32}$, and $\textbf{W}_{10}\in\mathbb{R}^{d \times 32}$ are weight matrices, and $\textbf{t}$ is the initialize temporal embeddings.
Afterward, the temporal embeddings are updated through a single-layer RGCN \cite{rgcn} to avoid over-smoothing:
\begin{equation}
    \textbf{t}_i^{\prime \prime}=\sigma(\sum_{r \in \mathcal{R}_{TG}} \frac{1}{\mathcal{N}_i} \textbf{W}_{r}\textbf{t}_j^{\prime}+\textbf{W}_{11}\textbf{t}_i^{\prime}) \textnormal{,}
\end{equation}
where $i,j\in \mathcal{T}$, $\sigma$ is the RReLU activation function, $\mathcal{N}_i$ is the number of nodes connected with $t_i$, $\textbf{W}_{r} \in\mathbb{R}^{d \times d}$ and $\textbf{W}_{11} \in\mathbb{R}^{d \times d}$ are the relation specific matrix and self-loop weight in RGCN.
\subsection{Indicator} \label{ind}
The indicator collects statistics on historical events and refining the scores \cite{tirgn} generated by Time-aware Decoders. Specifically, for each given query $q=(s,r,?,t)$, where $q \in \mathcal{Q}_t$, we construct an indicator matrix $\mathbf{I}_t^{s,r}\in \mathbb{Z}^{|\mathcal{E}| \times |\mathcal{R}| \times |\mathcal{E}|}$, which encompasses all facts occurring before timestamp $t$ with a subject entity $s$ and relation $r$. In this matrix, the values corresponding to objects are set to 1 if their events occurred before timestamp $t$, and 0 otherwise.

The primary purpose of the indicator matrix is to narrow down the predicted entity range. It achieves this by distinguishing between entities with historical relevance to the given query and those without. This selective focus enhances the precision and historical relevance of predictions within the $\mathsf{LMS}$.
\begin{table*}[]
\caption{Statistics of ICEWS14s, ICEWS18, ICEWS05-15, GDELT and ICEWS14 datasets.}
\resizebox{\textwidth}{!}{%
\begin{tabular}{cccccccc}
\hline
Dataset & Training Facts & Validation Facts & Testing Facts & Entities $\mathcal{E}$ & Relations $\mathcal{R}$ & Timestamps $\mathcal{T}$ & Time Granularity \\
\hline
ICEWS14s & 74,845 & 8,514 & 7,371 & 7,128 & 230 & 365 & 1 day \\
ICEWS18 & 373,018 & 45,995 & 49,545 & 23,033 & 256 & 304 & 1 day \\
ICEWS05-15 & 368,868 & 46,302 & 46,159 & 10,094 & 251 & 4,017 & 1 day \\
GDELT & 1,734,399 & 238,765 & 305,241 & 7,691 & 240 & 2976 & 15 mins \\
ICEWS14 & 63,685 & 13,823 & 13,222 & 7,128 & 230 & 365 & 1 day \\
\hline
\end{tabular}%
}
\label{dataset}
\end{table*}
\subsection{Time-aware Decoders} \label{dec}
We adopt two time-aware decoders to achieve comprehensive score prediction by leveraging historical and global raw information from TKGs. These decoders are built upon the ConvTransE framework \cite{convtranse} and are integrated with temporal information.
\subsubsection{Historical Decoder}
\begin{table*}[ht]
\caption{TKG entity extrapolation results on ICEWS14s, ICEWS18, and ICEWS05-15. The filtered MRR, H@1, H@3, and H@10 metrics are multiplied by 100. The best results are boldfaced and the second SOTAs are underlined. All baseline results are reported by previous works.}
\resizebox{\textwidth}{!}{%
\begin{tabular}{l|cccccccccccc}
\hline
\multirow{2}{*}{Model} & \multicolumn{4}{c}{ICEWS14s} & \multicolumn{4}{c}{ICEWS18} & \multicolumn{4}{c}{ICEWS05-15} \\ \cline{2-13} 
 & MRR & H@1 & H@3 & H@10 & MRR & H@1 & H@3 & H@10 & MRR & H@1 & H@3 & H@10 \\ \hline
DistMult & 15.44 & 10.91 & 17.24 & 23.92 & 11.51 & 7.03 & 12.87 & 20.86 & 17.95 & 13.12 & 20.71 & 29.32 \\
ComplEx & 32.54 & 23.43 & 36.13 & 50.73 & 22.94 & 15.19 & 27.05 & 42.11 & 32.63 & 24.01 & 37.50 & 52.81 \\
ConvE & 35.09 & 25.23 & 39.38 & 54.68 & 24.51 & 16.23 & 29.25 & 44.51 & 33.81 & 24.78 & 39.00 & 54.95 \\
ConvTransE & 33.80 & 25.40 & 38.54 & 53.99 & 22.11 & 13.94 & 26.44 & 42.28 & 33.03 & 24.15 & 38.07 & 54.32 \\
RotatE & 21.31 & 10.26 & 24.35 & 44.75 & 12.78 & 4.01 & 14.89 & 31.91 & 24.71 & 13.22 & 29.04 & 48.16 \\ \hline
RE-NET & 36.93 & 26.83 & 39.51 & 54.78 & 29.78 & 19.73 & 32.55 & 48.46 & 43.67 & 33.55 & 48.83 & 62.72 \\
CyGNet & 35.05 & 25.73 & 39.01 & 53.55 & 27.12 & 17.21 & 30.97 & 46.85 & 40.42 & 29.44 & 46.06 & 61.60 \\
xERTE & 40.02 & 32.06 & 44.63 & 56.17 & 29.31 & 21.03 & 33.51 & 46.48 & 46.62 & 37.84 & 52.31 & 63.92 \\
TITer & 40.97 & 32.28 & 45.45 & 57.10 & 29.98 & 22.05 & 33.46 & 44.83 & 47.60 & 38.29 & 52.74 & 64.86 \\
RE-GCN & 41.75 & 31.57 & 46.70 & 61.45 & 32.62 & 22.39 & 36.79 & 52.68 & 48.03 & 37.33 & 53.90 & 68.51 \\
EvoKG & 27.18 & - & 30.84 & 47.67 & 29.28 & - & 33.94 & 50.09 & - & - & - & -  \\
CEN & 43.34 & 33.18 & 48.49 & 62.58 & 32.66 & 22.55 & 36.81 & 52.50 & - & - & - & -  \\
TiRGN & 44.61 & 33.90 & 50.20 & 64.89 & 33.66 & 23.19 & 37.99 & 54.22 & 50.04 & 39.25 & 56.13 & 70.71 \\
RPC & - & - & - & - & \textbf{34.91} & \textbf{24.34} & \underline{38.74} & \textbf{55.89} & \underline{51.14} & \underline{39.47} & \underline{57.11} & \underline{71.75} \\
\hline
$\mathsf{LMS}$ & \textbf{46.76} & \textbf{36.12} & \textbf{51.97} & \textbf{67.29} & \underline{34.82} & \underline{24.20} & \textbf{39.30} & \underline{55.54} & \textbf{52.59} & \textbf{41.92} & \textbf{58.71} & \textbf{72.70} \\ \textit{improve}$\Delta$
 & 4.82\%$\uparrow$ & 6.55\%$\uparrow$ & 3.53\%$\uparrow$ & 3.70\%$\uparrow$ & 0.26\%$\downarrow$ & 0.58\%$\downarrow$ & 1.45\%$\uparrow$ & 0.63\%$\downarrow$ & 2.84\%$\uparrow$ & 6.21\%$\uparrow$ & 2.80\%$\uparrow$ & 1.32\%$\uparrow$  \\ \hline
\end{tabular}%
}
\label{exp1}
\end{table*}
The historical decoder leverages the indicator $\mathbf{I}$ to make predictions by considering their historical relevance:
\begin{equation}
    p_H(o|s, r, t, G_{t-1}) = softmax(\mathbf{ConvTransE}(\textbf{s},\textbf{r},\textbf{t}^{\prime \prime})\textbf{GE}^{\top}_t \odot \mathbf{I}_t^{s,r})\textnormal{,}
\end{equation}
where $\cdot^{\top}$ and $\odot$ denote transposition and element-wise multiplication, respectively. $\mathbf{I}_t^{s,r}$ is the indicator matrix corresponding to the prediction timestamp $t$.
\subsubsection{Raw Decoder}
The raw decoder, in contrast, does not incorporate the indicator matrix $\mathbf{I}_t^{s,r}$. It considers the information across the entire sliced sequence of TKGs from a global perspective, which is defined as:
\begin{equation}
    p_R(o|s, r, t, G_{t-1}) = softmax(\mathbf{ConvTransE}(\textbf{s},\textbf{r},\textbf{t}^{\prime \prime})\textbf{GE}^{\top}_t)\textnormal{.}
\end{equation}
These two scores are combined using a weight hyper-parameter $\alpha$. In this paper, we select an appropriate historical rate of $\alpha=0.3$, consistent with the previous baseline \cite{tirgn}, to obtain the final scores:
\begin{equation}
    p(o|s, r, t, G_{t-1}) = \alpha p_H(o|s, r, t, G_{t-1}) + (1-\alpha) p_R(o|s, r, t, G_{t-1})\textnormal{.}
\end{equation}
It is worth noting that the relation prediction is calculated similarly.
\subsection{Learning Objective}
$\mathsf{LMS}$ utilizes Cross-Entropy as the loss function. Notably, we consider optimizing the relation prediction to further improve the performance of entity prediction task similar to TiRGN \cite{tirgn} and HGLS \cite{hgls}. Meanwhile, a coefficient $\beta$ is set to control the weight of two tasks. Therefore, the learning objective is formalized as follows:
\begin{equation}
    \begin{aligned}
    \mathcal{L} &= \beta\sum_{(s,r,t)\in\mathcal{Q}^e_t}y^e_t \log p(o|s, r, t, G_{t-1}) \\
    &+ (1-\beta)\sum_{(s,o,t)\in\mathcal{Q}^r_t}y^r_t \log p(r|s, o, t, G_{t-1})\textnormal{,}
    \end{aligned}
\end{equation}
where $\mathcal{Q}^e_t$ and $\mathcal{Q}^r_t$ denotes the queries of entities and relations at $t$, $y^e_t$ and $y^r_t$ is the truth label vectors for the two tasks.
\section{Experiments}
In this section, we perform extensive experiments on five event-based TKG datasets to demonstrate the state-of-the-art performances of our model, which can be summarized in the following questions.
\begin{itemize}[leftmargin=*]
\item \textbf{Q1}: How does $\mathsf{LMS}$ perform compared with state-of-the-art TKG extrapolation approaches on the entity prediction task?
\item \textbf{Q2}: How do the evolutional, union, and temporal graphs from TKGs affect the performance of $\mathsf{LMS}$?
\item \textbf{Q3}: How do temporal information integration and adaptive gate components affect the performance of $\mathsf{LMS}$?
\item \textbf{Q4}: How does the performance fluctuation of $\mathsf{LMS}$ with different settings of hyper-parameters?
\end{itemize}
\subsection{Experimental Setup}
\subsubsection{Datasets.}
This paper leverages five real-world event-based TKGs that have been extensively utilized in prior research. Among them, four pertain to the Integrated Crisis Early Warning System (ICEWS) \cite{icews}: ICEWS14s, ICEWS18, ICEWS05-15, ICEWS14. Besides, for comprehensive comparison and to further demonstrate the robustness of $\mathsf{LMS}$, the Global Database of Events, Language, and Tone (GDELT) \cite{gdelt} dataset is also incorporated.
Consistent with earlier works \cite{regcn,tirgn,rpc}, the datasets undergo partitioning into training, validation, and testing subsets based on their timestamps. Further details regarding the statistical attributes of these datasets are presented in Appendix \ref{DBS}.
Notably, in order to compare with different state-of-the-art methods, we use two different ICEWS14 datasets, distinguished by whether they end with an s or not.
The statistics of five TKG datasets in this paper are shown in Table \ref{dataset}.
\subsubsection{Baselines.}
We compare our proposed $\mathsf{LMS}$ with static KG reasoning methods \cite{distmult,complex,conve,convtranse,rotate} and the state-of-the-art TKG extrapolation models, including RE-NET \cite{renet}, CyGNet \cite{cygnet}, xERTE \cite{xerte}, TITer \cite{titer}, RE-GCN \cite{regcn}, EvoKG \cite{evokg}, CEN \cite{cen}, TiRGN \cite{tirgn}, RPC \cite{rpc}, and HGLS \cite{hgls}. The details of these extrapolation methods are shown in the following:
\begin{itemize}[leftmargin=*]
\item RE-NET \cite{renet} addresses the challenge of extrapolation, utilizing a combination of graph-based and sequential modeling techniques to capture both temporal and structural interactions among entities.
\item CyGNet \cite{cygnet} introduces a copy-generation mechanism and constructs a historical vocabulary associated with past events, amplifying the significance of these historical facts.
\item xERTE \cite{xerte} proposes a time-aware GAT that leverages temporal relations to explore subgraphs using temporal information, facilitating the prediction of ultimate outcomes.
\item TITer \cite{titer} integrates reinforcement learning into TKG reasoning and introduces a search strategy to identify pertinent historical facts aligned with user queries.
\item RE-GCN \cite{regcn} incorporates a recurrent unit based on GNN to capture both the structural and temporal characteristics of TKGs. Furthermore, it devises a static graph to augment initial embeddings specifically for the ICEWS datasets.
\item EvoKG \cite{evokg} captures the dynamic structural and temporal patterns inherent in TKGs by employing recurrent event modeling.
\item CEN \cite{cen} employs a length-aware CNN within a curriculum learning framework, while also introducing an online configuration to dynamically accommodate shifts in evolutional patterns.
\item TiRGN \cite{tirgn} proposes time-aware ConvTransE and ConvTransR decoders and incorporates a global historical vocabulary to account for the presence of entities or relations in previous contexts.
\item RPC \cite{rpc} mines the intra-snapshot graph structural correlations and periodic inter-snapshot temporal interactions via two correspondence units.
\item HGLS \cite{hgls} designs a hierarchical relational GNN to capture correlations within concurrent facts, and temporal dependencies across snapshots at a global graph level.
\end{itemize}
\subsubsection{Implementation Details.}
The initial temporal embedding dimension is configured at 32, while the remaining vectors are set to 200. A dropout rate of 0.2 is applied uniformly across all modules. The layers of graph aggregators for the evolutional, union, and temporal graphs are set to 2, 2, and 1, respectively. Additionally, we utilize the Adam optimizer with a learning rate parameterized to 0.001.
For the optimal evolutional and union length $k$, we conduct grid research according to the validation results, and the best value of $k$ is set to 25, 11, 25, 45, and 17 for ICEWS14s, ICEWS18, ICEWS05-15, GDELT, and ICEWS14, respectively. Noted that, to fair comparison, the historical rate $\alpha$ and task coefficient $\beta$ are set to 0.3 and 0.7 separately, and the static graph learning module is employed on ICEWS datasets similar to our baselines \cite{regcn,tirgn,rpc}.
\subsubsection{Evaluation Metrics.}
In the experiments, we utilize the widely-used time-aware filtered metrics \cite{xerte,tirgn,rpc} MRR and Hits@{1,3,10}.
Additionally, to demonstrate a wider outperform of $\mathsf{LMS}$, we report a part of experiments under raw setting \cite{regcn,hgls} and remove the above static module in comparison with HGLS.
\begin{table}[]
\centering
\caption{TKG entity extrapolation results on GDELT and ICEWS14 with time-aware filtered metrics.}
\resizebox{\linewidth}{!}{%
\begin{tabular}{l|cccccccc}
\hline
\multirow{2}{*}{Model}  & \multicolumn{4}{c}{GDELT} & \multicolumn{4}{c}{ICEWS14} \\ \cline{2-9}  & MRR & H@1 & H@3 & H@10 & MRR & H@1 & H@3 & H@10 \\ 
\hline
RE-NET & 19.55 & 12.38 & 20.80 & 34.00 & 39.86 & 30.11 & 44.02 & 58.21 \\
CyGNet & 20.22 & 12.35 & 21.66 & 35.82 & 37.65 & 27.43 & 42.63 & 57.90 \\
xERTE & 19.45 & 11.92 & 20.84 & 34.18 & 40.79 & 32.70 & 45.67 & 57.30 \\
TITer & 18.19 & 11.52 & 19.20 & 31.00 & 41.73 & 32.74 & 46.46 & 58.44\\
RE-GCN & 19.69 & 12.46 & 20.93 & 33.81 & 42.00 & 31.63 & 47.20 & 61.65 \\
EvoKG & 19.28 & - & 20.55 & 34.44 & 27.18 & - & 30.84 & 47.67 \\
CEN & 21.16 & 13.43 & 22.71 & 36.38 & 44.61 & 33.90 & 50.20 & 64.89 \\
TiRGN & 21.67 & 13.63 & 23.27 & 37.60 & 43.81 & 33.49 & 48.90 & 63.50 \\
RPC & \underline{22.41} & \underline{14.42} & \underline{24.36} & \underline{38.33} & \underline{44.55} & \underline{34.87} & \underline{49.80} & \underline{65.08} \\
\hline
$\mathsf{LMS}$ & \textbf{22.94} & \textbf{14.49} & \textbf{24.80} & \textbf{39.66} & \textbf{45.98} & \textbf{35.77} & \textbf{51.12} & \textbf{65.91} \\
\textit{improve}$\Delta$ & 2.37\%$\uparrow$ & 0.49\%$\uparrow$ & 1.81\%$\uparrow$ & 3.47\%$\uparrow$ & 3.21\%$\uparrow$ & 2.58\%$\uparrow$ & 2.65\%$\uparrow$ & 1.28\%$\uparrow$ \\ 
\hline
\end{tabular}%
}
\label{exp2}
\end{table}
\subsection{Results (RQ1)}
\begin{figure}[htbp]
    \begin{minipage}[h]{0.47\linewidth}
        \includegraphics[width=\linewidth]{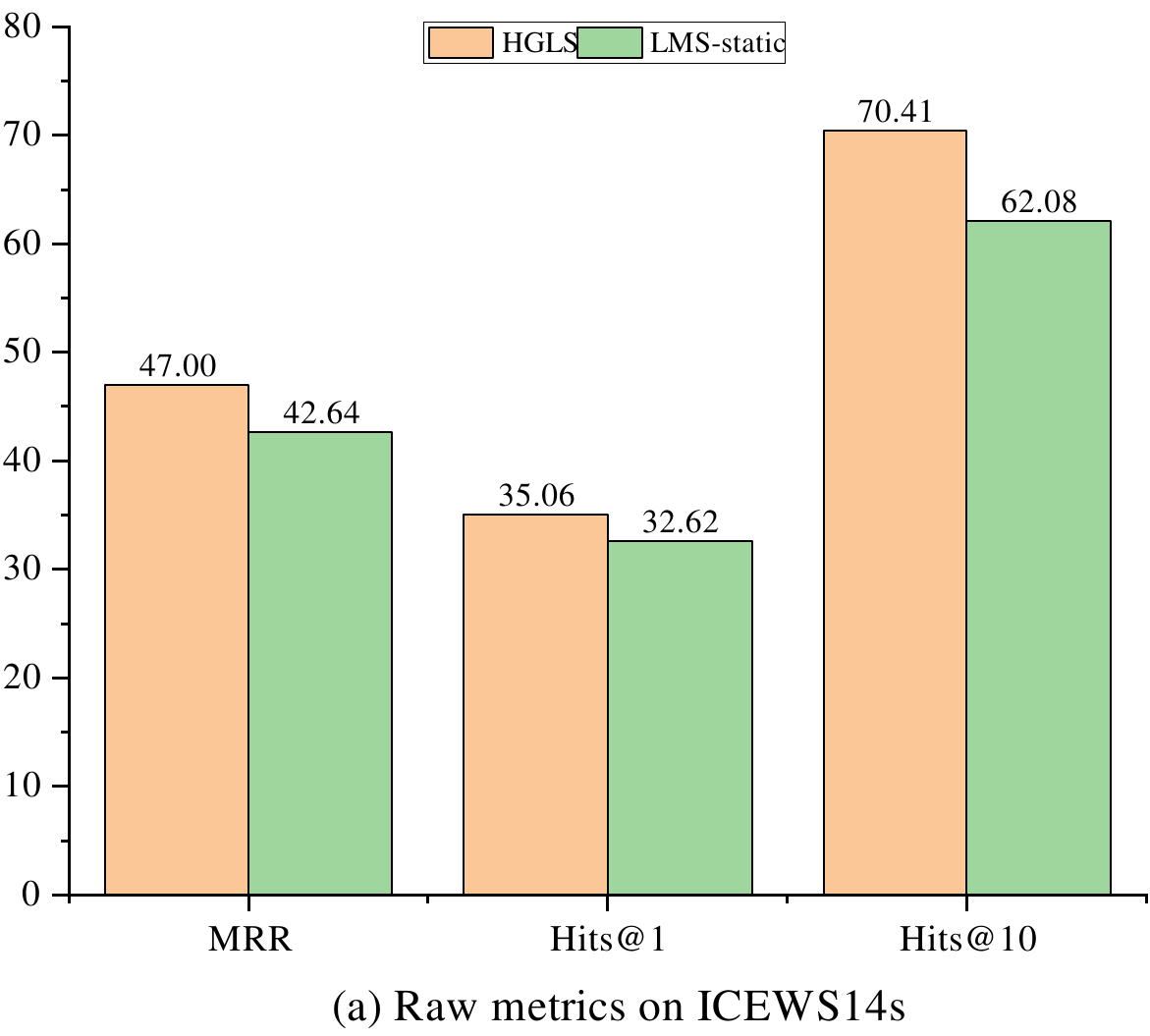}
    \end{minipage}%
    \begin{minipage}[h]{0.47\linewidth}
        \centering
        \includegraphics[width=\linewidth]{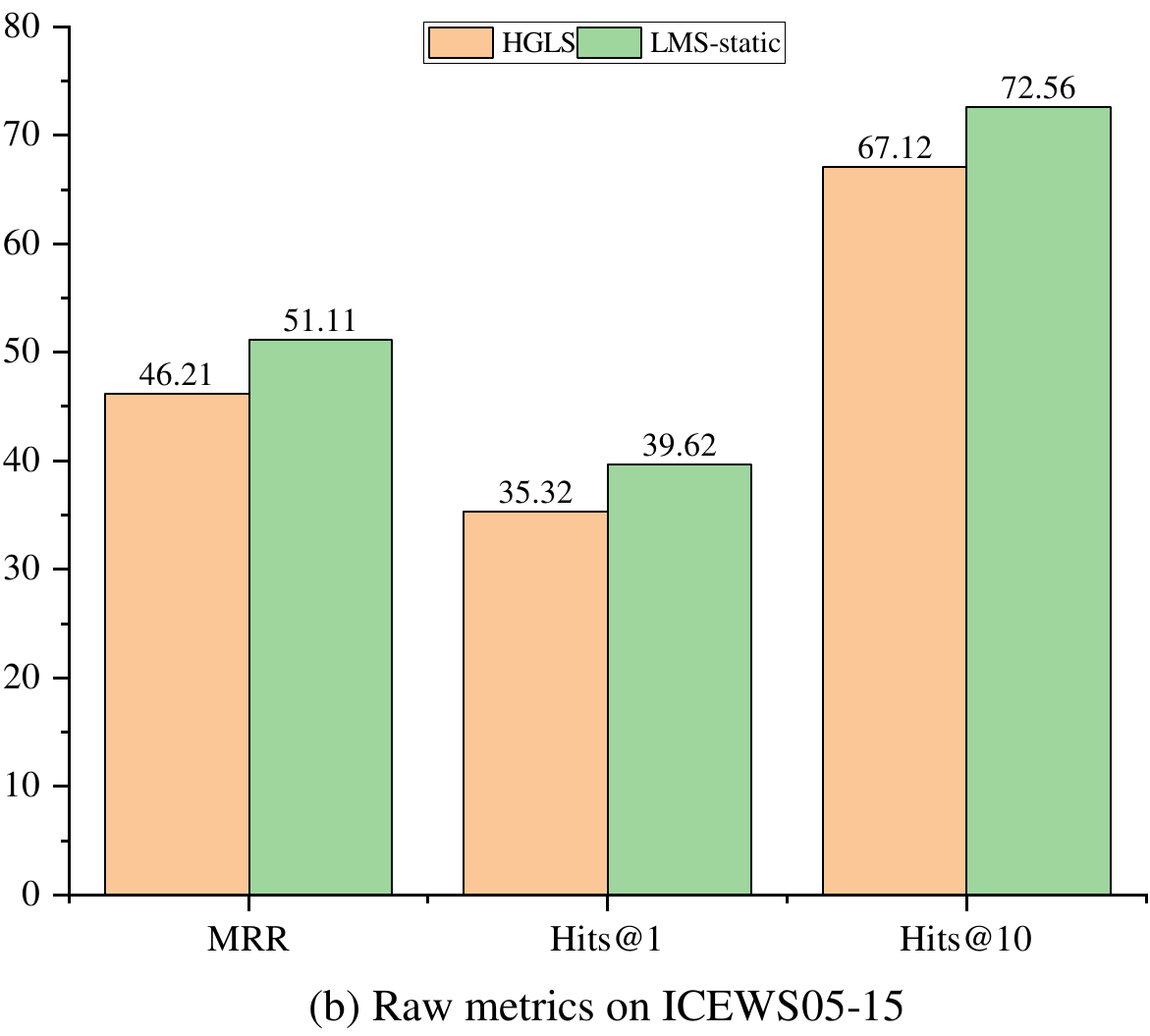}
    \end{minipage}
    \\
    \begin{minipage}[h]{0.47\linewidth}
        \centering
        \includegraphics[width=\linewidth]{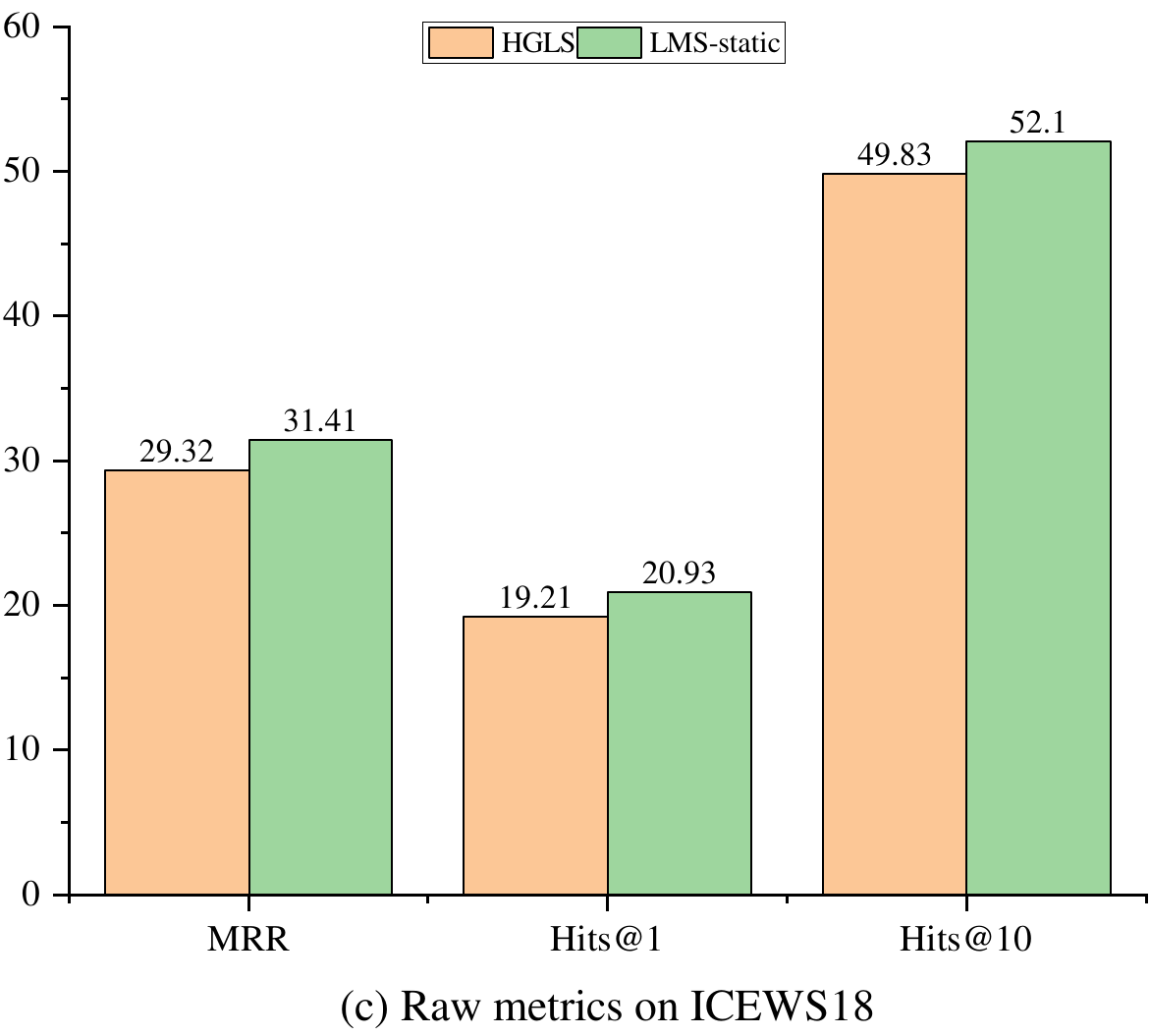}
    \end{minipage}
    \begin{minipage}[h]{0.47\linewidth}
        \centering
        \includegraphics[width=\linewidth]{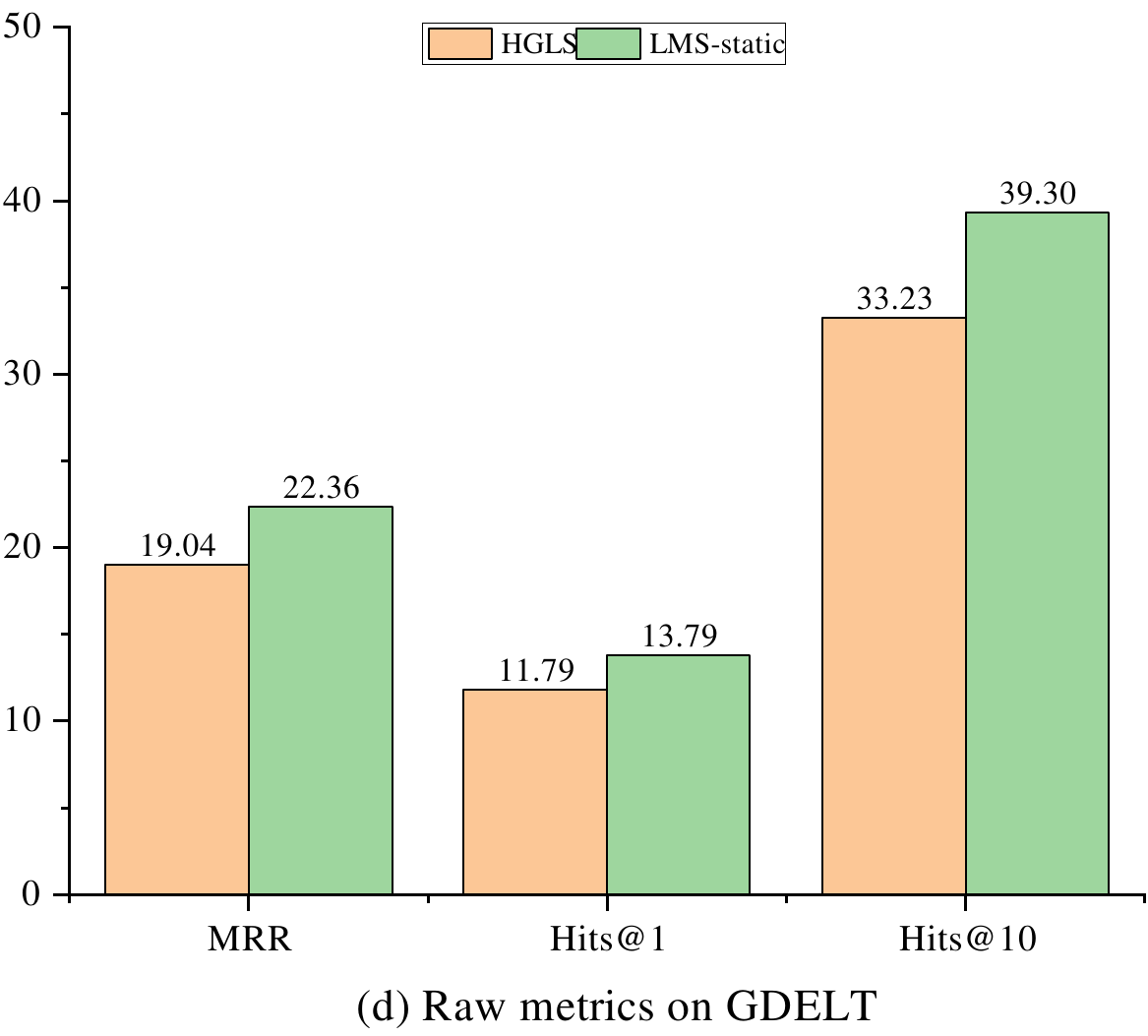}
    \end{minipage}
    \caption{TKG entity extrapolation results in comparison with HGLS under raw setting on ICEWS14s, ICEWS05-15, ICEWS18, and GDELT.}
    \label{hgls}
\end{figure}

The performances of $\mathsf{LMS}$ and other advanced methods on entity prediction task are shown in Table \ref{exp1}, \ref{exp2}, and Figure \ref{hgls}. 
Compared with the static KG reasoning methods (e.g., ConvE, RotatE), the previous TKG extrapolation models and $\mathsf{LMS}$ significantly outperform on all benchmarks, affirming the importance of temporal information in modeling TKG features and future prediction. Furthermore, when compared with various existing TKG extrapolation methods, $\mathsf{LMS}$ emerges as the state-of-the-art in most scenarios with a significant average improvement of 2.60\% and 6.47\% on time-filtered and raw MRR across event-based benchmarks. These results provide valuable insights and compelling evidence addressing Q1, highlighting the superiority of $\mathsf{LMS}$.

Specifically, RE-NET only integrates limited neighboring semantics of TKGs, while CyGNet primarily focuses on capturing repetitive historical events but tends to overlook structural dependencies among snapshots. xERTE and TITer are constrained by the number of hops allowed for subgraph search. In contrast, $\mathsf{LMS}$ achieves superior performance with an average 15.41\% improvement on MRR by considering recent historical structural and temporal dependencies within TKGs. 

In comparison to a series of recent approaches \cite{regcn,evokg,cen,tirgn,rpc} that mainly model the most recent historical snapshots for TKG reasoning, $\mathsf{LMS}$ introduces a novel perspective that exclusively concentrates on events closely related to queries, and learning correlations across snapshots. RPC is the most advanced method in the above series, which introduces two correspondence units to capture structural and temporal dependencies from relation and snapshot views, but it still lacks the aspect of explicitly modeling periodic semantics and structural correlations for time vectors. Consequently, the MRR and Hits@1 of $\mathsf{LMS}$ outperform RPC by 2.04\% and 2.18\% on average, indicating the significance of temporal semantics for future prediction. 

Moreover, as depicted in Figure \ref{hgls}, $\mathsf{LMS}$ surpasses HGLS, which is adept at learning long-term representations for TKG reasoning, with an average improvement over 6.47\% and 7.78\% on raw MRR and Hits@1 across four datasets. This outstanding stems from considering a broader range of TKG features in $\mathsf{LMS}$, including global historical statistics and timestamp dependencies. Meanwhile, in contrast to the global graph proposed in HGLS, the union graph in $\mathsf{LMS}$ can focus more precisely on key facts or entities, allowing for semantically rich representations while reducing the number of nodes and computational consumption.

It is worth noting that previous novel methods perform poorly on the GDELT dataset due to its fine-grained timestamps and the inclusion of abstract conceptual entities (e.g., government entity without additional annotations makes country distinctions challenging). In contrast, $\mathsf{LMS}$ has a significant improvement of 2.37\% and 17.44\% on GDELT across time-aware filtered and raw MRR, respectively, compared to state-of-the-art methods \cite{rpc,hgls}. This highlights the capability of $\mathsf{LMS}$ to effectively capture event correlations across different timestamps through the perspective of multi-graph learning. Additionally, $\mathsf{LMS}$ achieves these results without requiring extensive historical information.
\subsection{Ablation Studies (RQ2 and RQ3)}
To investigate the impact of the EGL, UGL, and TGL modules within $\mathsf{LMS}$ (Q2), as well as the different variants of entity or temporal representation fusion components for $\mathsf{LMS}$ performance (Q3), we conduct a series of ablation studies on ICEWS14s dataset. The findings are summarized in Table \ref{ablation} and are as follows:
\begin{itemize}[leftmargin=*]
    \item In the investigation of impaction with different graph learning modules, $\mathsf{LMS}$-EGL denotes that the final entity representations are based on the outputs of UGL, highlighting the significance of capturing evolutional patterns in TKGs. Conversely, $\mathsf{LMS}$-UGL removes the UGL that models key events directly related to queries but solely relies on EGL for entity representations. It emphasizes that UGL is able to effectively mine cross-temporal entity connections in recent historical TKGs and assess entity impact over different times. Additionally, $\mathsf{LMS}$+UGL(Entirety) replaces the query-specific union graph with an entire union graph containing all recently occurred facts. This modification leads to a 5.65\% drop in MRR, highlighting the importance of reducing the density of neighboring entities within the union graph. $\mathsf{LMS}$-TGL indicates that timestamp representations are solely derived from Time2Vec, showcasing the effectiveness of TGL in capturing correlations and periodic dependencies among timestamps.
    \item In the comparison of different components, $\mathsf{LMS}$-T(UGL) denotes that we remove the representation of time information in the graph aggregator of UGL, $\mathsf{LMS}$-T(Decoder) means that the time-aware decoder is replaced with normal ConvTransE without the above temporal representation. $\mathsf{LMS}$-GATE+Sum and $\mathsf{LMS}$-GATE+Linear represent two alternatives to the adaptive gate, which are replaced with average summation and linear transformation, respectively. Compared with the original $\mathsf{LMS}$, the MRR of $\mathsf{LMS}$-T(UGL) and $\mathsf{LMS}$-T(Decoder) decrease by 0.46 and 0.35 respectively, which illustrates the effectiveness of fusing temporal information in different components and also confirms that efficient representation requires more holistic utilization of temporal information. Meanwhile, $\mathsf{LMS}$-GATE+Sum and $\mathsf{LMS}$-GATE+Linear demonstrate the importance of adaptively integrating evolutional and union representations.
\end{itemize}
\begin{table}[h]
\centering
\caption{Ablation studies with filtered metrics on ICEWS14s.}
\resizebox{1\linewidth}{!}{%
\begin{tabular}{lcccc}
\hline
Model & MRR & H@1 & H@3 & H@10 \\
\hline
$\mathsf{LMS}$ (baseline) & \textbf{46.76} & 36.12 & \textbf{51.97} & \textbf{67.29} \\
\hline
$\mathsf{LMS}$-EGL & 43.41 & 33.36 & 48.15 & 63.10 \\
$\mathsf{LMS}$-UGL & 44.48 & 33.93 & 49.60 & 64.96 \\
$\mathsf{LMS}$+UGL(Entirety) & 44.12 & 33.79 & 49.13 & 64.20 \\
$\mathsf{LMS}$-TGL & 46.26 & 35.67 & 51.49 & 66.67 \\
\hline
$\mathsf{LMS}$-T(UGL) & 46.30 & 35.72 & 51.43 & 66.98 \\
$\mathsf{LMS}$-T(Decoder) & 46.39 & 35.67 & 51.72 & 66.70 \\
$\mathsf{LMS}$-GATE+Sum & 46.59 & \textbf{36.19} & 51.75 & 66.84 \\
$\mathsf{LMS}$-GATE+Linear & 45.36 & 34.82 & 50.69 & 65.78 \\
\hline
\end{tabular}%
}
\label{ablation}
\end{table}
\subsection{Hyper-parameter Sensitivity Analysis (RQ4)}
To further explore the sensitivity of $\mathsf{LMS}$ with hyper-parameters (Q3), we conduct investigations into the impact of the evolutional and union length $k$ and historical rate $\alpha$ on the performance of $\mathsf{LMS}$, which is shown in Figure \ref{hyp-analysis}.

\subsubsection{Effect of evolutional and union length $k$}
\begin{figure}[h]
\centerline{\includegraphics[width=0.95\linewidth]{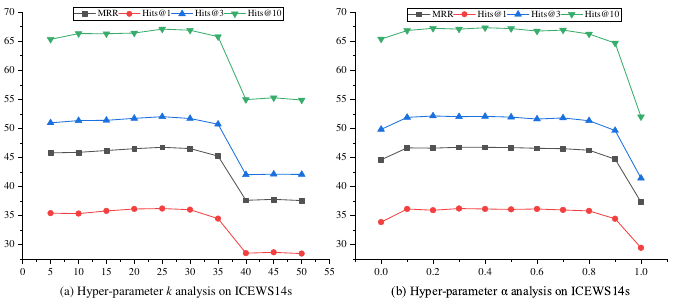}}
\caption{Analysis of hyper-parameter $k$ and $\alpha$ on ICEWS14s.}
\label{hyp-analysis}
\end{figure}

The hyper-parameter $k$ plays a pivotal role in controlling the historical length of both EGL and UGL simultaneously, thereby determining the number of evolutional iterations and connections within the union graph. As illustrated in Figure \ref{hyp-analysis}(a), $\mathsf{LMS}$ exhibits a compelling performance with various values of $k$. It surpasses HGLS, which employs the entire historical snapshots or an extensive length (e.g., 500) as a global graph. Notably, the performance rapidly decreases as $k$ ranges from 35 to 40, indicating that $\mathsf{LMS}$ does not heavily rely on an extensive historical context for effective extrapolation. It is worth noting that $\mathsf{LMS}$ maintains strong performance even with a smaller $k$ value, demonstrating its robustness and effectiveness. These findings highlight the ability of $\mathsf{LMS}$ to adapt to varying degrees of historical information, avoiding potential issues associated with overly dense neighbor structures that can impede the identification of crucial entities or facts.
\subsubsection{Effect of historical rate $\alpha$}
We assess the impact of the hyper-parameter $\alpha$, which governs the balance between the historical and raw decoder, over a range of values from 0.0 to 1.0. The phenomenon of $\alpha$ in $\mathsf{LMS}$ exhibits similarities to what was reported in TiRGN. Specifically, it demonstrates that the raw information carries a higher weight than historical facts pertaining to subject-relation pairs. Furthermore, it is noteworthy that LMS can still achieve impressive performance metrics even when $\alpha$ is set to 0.0. This suggests that $\mathsf{LMS}$ excels in encoding TKG features and is less dependent on historical repetitive patterns that can lead to discrepancies between the two decoders. This further demonstrates that UGL does not rely on pair constraints of relations and can capture a more comprehensive set of relevant facts compared to the historical indicator. On the contrary, when $\alpha=1.0$, the scores rely entirely on the historical indicator, the results become worse.
\section{Conclusion}
This paper introduces an innovative method $\mathsf{LMS}$ for TKG extrapolation, which excels at learning diverse TKG features from multi-graph perspectives, encompassing concurrent correlations and snapshot evolution, event dependencies across snapshots through a union graph, and realistic time semantics within a temporal graph. Furthermore, it incorporates various components for the fusion of comprehensive TKG information and precise prediction, including a time-aware attention mechanism, adaptive gate, and historical indicator. The experimental results on five event-based TKG datasets and various studies demonstrate the effectiveness and state-of-the-art performances of $\mathsf{LMS}$. In future research, we aim to develop a strategy that can dynamically adapt the history length for both the evolutional graph and union graph, and tailor periods of temporal graph to specific datasets.
\begin{acks}
\end{acks}

\bibliographystyle{ACM-Reference-Format}
\bibliography{manuscript}

\appendix

\end{document}